\documentclass[letterpaper, 10 pt, conference]{ieeeconf}  
\usepackage{graphicx}
 \usepackage{url} 

\IEEEoverridecommandlockouts                              % This command is only needed if 
\title{\LARGE \bf
\textit{myEye2Wheeler}: A Two-Wheeler Indian Driver Real-World Eye-Tracking Dataset
}

\author{Bhaiya Vaibhaw Kumar, Deepti Rawat, Tanvi Kandalla, Aarnav Nagariya, Kavita Vemuri% <-        
}

\begin{document}

\maketitle
\thispagestyle{empty}
\pagestyle{empty}

%%%%%%%%%%%%%%%%%%%%%%%%%%%%%%%%%%%%%%%%%%%%%%%%%%%%%%%%%%%%%%%%%%%%%%%%%%%%%%%%
\begin{abstract}

This paper presents the \textit{myEye2Wheeler} dataset, a unique resource of real-world gaze behaviour of two-wheeler drivers navigating complex Indian traffic. Most datasets are from four-wheeler drivers on well-planned roads and homogeneous traffic. Our dataset offers a critical lens into the unique visual attention patterns and insights into the decision-making of Indian two-wheeler drivers. The analysis demonstrates that existing saliency models, like TASED-Net, perform less effectively on the \textit{myEye-2Wheeler} dataset compared to when applied on the European 4-wheeler eye tracking datasets (DR(Eye)VE), highlighting the need for models specifically tailored to the traffic conditions. By introducing the dataset, we not only fill a significant gap in two-wheeler driver behaviour research in India but also emphasise the critical need for developing context-specific saliency models. The larger aim is to improve road safety for two-wheeler users and lane-planning to support a cost-effective mode of transport. 

\end{abstract}

%%%%%%%%%%%%%%%%%%%%%%%%%%%%%%%%%%%%%%%%%%%%%%%%%%%%%%%%%%%%%%%%%%%%%%%%%%%%%%%%
\section{INTRODUCTION}

Traffic safety is a global concern, with road accidents causing significant human and economic losses annually \cite{who1}. Two-wheeler drivers in countries like India are particularly vulnerable, navigating complex traffic environments \cite{ankit2}. Driver assistance technologies have improved safety, relying on understanding driver gaze behaviour \cite{Sharma_review3}. However, research on two-wheelers, especially in chaotic traffic like India's, is lacking \cite{khan4,Kotseruba5}. The drivers in India face challenges \cite{khan4} like dense traffic and unpredictable interactions like pedestrians and animals entering the lanes, sudden lane changes by other vehicles without indicators, wrong-side overtakes, vehicles of different engine capacity, un-synchronised traffic lights, potholes, etc. The physics and mechanics of driving and the driver-external world interaction are different from a 4-wheeler, in particular, the higher risk to life in an accident. Given the complexity, We advocate for specialised datasets to develop tailored Advanced Driver-Assistance Systems (ADAS).

Existing datasets like \textit{DR(eye)VE} and \textit{Look Both Ways} \cite{alletto2016dr6,Isaac7} cater to four-wheeler drivers and thus fail to capture situations faced by Indian two-wheeler drivers and the decisions made by them, such as choosing to move through narrow gaps between other vehicles, frequently changing lanes, overtaking from the wrong side, all while also being aware of the aforementioned unpredictability of Indian roads. The \textit{myEye2Wheeler} dataset addresses this gap.

\subsection{Related Works}
\subsubsection{Open Source Driver Gaze Datasets} 
While many studies rely on simulators for data collection \cite{fisher2007empirical8,deng2019prediction9}, advancements in wearable eye trackers and vision processing algorithms have enabled real-world eye-tracking. This technology has diverse applications, such as investigating distractions from mobile usage \cite{ojstersek2019eye10}, the effects of stress on driving behaviour \cite{wang2023eye11}, and assessing the efficacy of driver assistance systems \cite{said2018real12}.

Most driver eye-tracking research focuses on four-wheelers, with limited attention to two-wheelers \cite{distasi2011behavioral13,papakostopoulos2020semantic14,hosking2010visual15}, outside Western countries. Developing robust datasets is crucial for refining prediction algorithms derived from ground-truth gaze estimation. This section reviews notable datasets to emphasise the unique contributions of the \textit{myEye2Wheeler} dataset.
\subsubsection{Gaze Zone Predictor datasets}
The DG-UNICAMP dataset,  by Ribeiro et al. \cite{ribeiro16}, integrates RGB, IR, and depth cameras, but its stationary vehicle setup limits capturing dynamic driving scenarios.
The Driver Gaze in the Wild (DGW) Dataset, introduced by Ghosh et al. \cite{ghosh17}, improves representation with a large subject pool but retains the limitations of a stationary setup.
The Driver Monitoring Dataset (DMD) \cite{ortega18} focuses on driver behaviour and gaze zone classification of interior components of the car and, similar to previous datasets, does not include gaze dynamics to the external traffic or road.

\subsubsection{Traffic Driver Gaze Datasets}
The \textit{DADA-2000} dataset \cite{dada19} is a benchmark for driver attention prediction in accident scenarios, comprising 2000 video sequences with detailed annotations covering diverse driving conditions and accident types. It provides fixation maps, saccade scan paths, accident categories, and spatial crash object locations, offering insights into the relationship between driver attention and accidents.

The \textit{DR(eye)VE }dataset \cite{alletto2016dr6} collected on European roads focuses on driver gaze research in varied lighting conditions. It includes 74 videos captured from a wearable eye tracker and a roof-mounted camera, providing synchronised gaze coordinates for each frame.

The \textit{Look Both Ways} (LBW) Dataset \cite{Isaac7} offers gaze data from a wearable eye tracker and dashboard-mounted camera images. It provides insights into gaze patterns and analysis derived from the face images of drivers.

The \textit{DGAZE} dataset \cite{Dua20} captures traffic gaze data in Indian traffic conditions, featuring face and traffic images with gaze annotations, albeit collected in a lab setting mimicking 4-wheeler driving conditions.

\subsection{Research Gap}

The majority of driver behavioural research concentrates on Western countries, where four-wheelers dominate, and two-wheelers are not the primary mode of transportation \cite{2_3_wheelers21}. In these settings, traffic also tends to be more uniform compared to the conditions on Indian roads. Additionally, the driving approach in Western countries differs, as there is less traffic and a disciplined approach to lane switching or overtaking, leading to an almost automatic motor-memory, cognitive processing and attention allocation by drivers \cite{transweb22,di202023}. 

Research on real-world eye-tracking studies involving two-wheelers is nascent, primarily for vehicle flow-behavior analysis rather than assistive systems design or urban road planning. But a thorough analysis of the basis for strategic shifts in visual gaze and the decision-making factors, such as estimation of surround vehicle engine capacity, size of surround vehicles, the experience-based prediction of the other vehicle driver behaviour, traffic density, and drivers' self-discipline, can extend the insights to better inter-vehicle interaction and road design.

Our study aims to achieve four main goals: a) curating an eye-tracking real-world dataset for further research and analysis, b) extracting driver gaze behaviour, and c) developing a saliency model. This paper focuses on the initial step of curating a dataset to advance research in the field and support the accomplishment of the remaining goals.

\section{DATA ACQUISITION}
\begin{figure} \centering
\includegraphics[scale = 0.7]{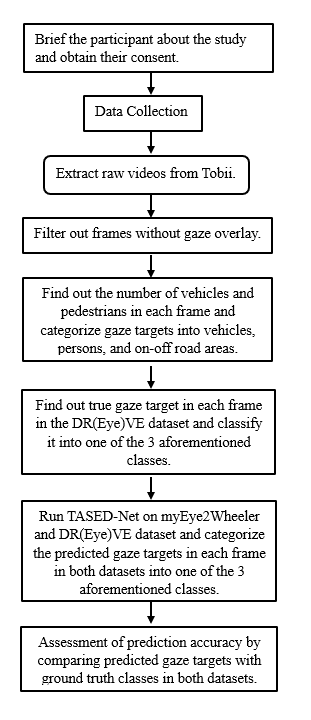}
\caption{Flowchart representing the key steps in this study} \label{flowch}
\end{figure}

\subsection{Participants}
%A total of 51 participants participated in the study (age
% range: 19-50 years; male = 44, female = 7). Data from
% 11 individuals (male = 10, female = 1) were excluded due
% to synchronisation errors with the Tobii Glasses controller
% app, resulting in gaze data loss. 
A total of 40 participants volunteered for the study (age range: 19-50 years; male = 34, female = 6). Participants used either their personal two-wheelers or were provided with a TVS Jupiter, a gearless scooter with a 109.7 cc single-cylinder air-cooled engine. All participants held valid two-wheeler driving licenses issued by the Regional Transport Office authority.

Participants were categorised into two groups based on driving experience: `experienced drivers' with over a decade of 2-wheeler driving and `novice drivers' with 1 to 6 years of experience since obtaining a license. The dataset comprises 21 novice participants (age: 19-25 years; male = 18, female = 3) and 20 experienced participants (age: 27-50 years; male = 16, female = 4).

Safety measures were rigorously enforced during the study. Participants were instructed to adhere to a speed limit of 60 km/h and follow all traffic rules. Both participants and the experimenter wore helmets throughout the ride. Participants were assured of anonymity and that their data would only be used for research. A nominal compensation of INR 250 was provided to acknowledge their time and effort.
 
 \subsubsection{Ethics.} In compliance with ethical standards, all participants provided informed consent, acknowledging the voluntary nature of their involvement, and the research protocol received approval from the Institute Ethics Board.

\subsection{Wearable Eye Tracker}

\begin{figure*} \centering
\includegraphics[scale = 0.27]{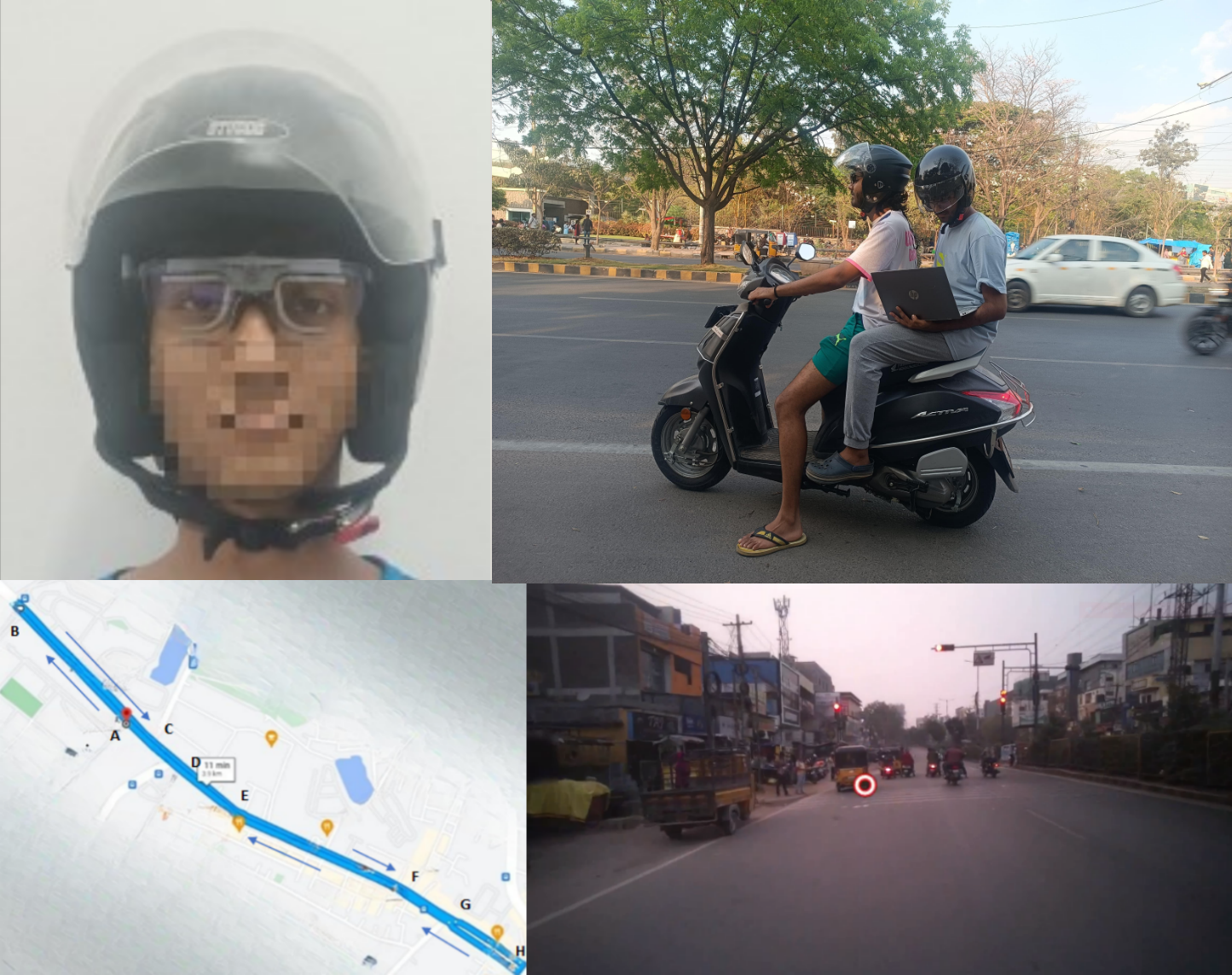}
\caption{In clockwise order from top-left: (a) Participant wearing the eye tracker inside his helmet and, (b) Experiment setup while on the road and, (c) Point G on the route, and (d) Top view of the route described where arrows denote the direction of traffic} \label{fig2}
\end{figure*}

The eye tracker used to collect data is the Tobii Pro Glasses 2. It features an advanced wearable Eye-Tracking Unit, with a sampling rate of 50Hz set for this study. The wearable unit (Figure \ref{fig2}a) includes a front-scene camera which captures video at a resolution of 1920 × 1080 at 25 fps in the H.264 format. With a wide field of view (90 deg. in 16:9 aspect ratio), the camera comprehensively captures the participant's ego-centric view, offering horizontal and vertical FOV at approximately 82 degrees and 52 degrees, respectively. Four IR cameras are placed inside the glasses and facing the eye to capture eye movements. Gyroscope and accelerometer sensors contribute to the device's stability and accuracy. 

\subsubsection{Pre-Experiment Briefing.} Participants received a thorough orientation session on the study's objectives and the predetermined route. They were instructed to wear helmets and provided with a wearable eye-tracking device. Meticulous calibration of the eye-tracker was conducted for each participant to ensure accurate data collection. A 1.5km test ride in a controlled route (inside the college campus) was included to address potential concerns or discomfort associated with wearing the tracking glasses and to assess driving capability.

\subsubsection{Route.}
%The selected route covers a total distance of 3.9 kilometres (Figure \ref{fig2}d), following a round-trip path along a busy urban road (start lat and long coordinates : 
% 17.445052020270055, 78.35272802371485; end: 17.438703326713025, 78.36373697796877; same return route)
%  in Hyderabad, India.
The selected route covers a total distance of 3.9 kilometres (Figure \ref{fig2}d), following a round-trip path along a busy urban road 
in Hyderabad, India. This well-maintained road with minimal potholes features lane markings, traffic signals, zebra crossings, and U-turns. The route includes a pedestrian walking space (paved partially) with pedestrian crossings. The 2-way four lanes (mostly) have a divider to ensure no crossing by pedestrians except in the designated zebra markings and a timer controlled with a signal. Data collection occurred between 12:00 PM and 05:00 PM to avoid office rush hours and high jams while also capturing a steady traffic flow.

Key points of the experimental route - 
\begin{enumerate}
    \item Participants start at point \textbf{A} (lat/long:17.446047, 78.351490) on the left side of the road and progress towards point \textbf{B} ( 17.449355, 78.348346), initially facing the sun, and execute a U-turn.
    % \item Progress towards point \textbf{B}, initially facing the sun, and execute a U-turn.
    \item They reach point \textbf{C} (17.445530, 78.352351), an intersection with left-turning traffic and vehicles from that road taking right turns onto the route and approaching a traffic signal at point \textbf{D} (17.445104, 78.352853), experiencing potential congestion during red signal intervals.
    % \item Approach a traffic signal at point \textbf{D}, experiencing potential congestion during red signal intervals.
    % \item A divider after \textbf{D}, restricting pedestrian crossings.
    \item They proceed approximately 200 meters to point \textbf{E} (17.443797, 78.354507), marking a U-turn, but continue straight, navigating narrow left turns, passing petrol pumps, and encountering a pedestrian-friendly traffic signal at point \textbf{F} (17.441076, 78.360163). 
    % \item Move from \textbf{E} to \textbf{F}, navigating narrow left turns, passing petrol pumps, and encountering a pedestrian-friendly traffic signal at \textbf{F}.
    \item Around 200 meters from \textbf{F}, they arrive at point \textbf{G} (17.440487, 78.361174), where the road splits with a right leading to a flyover and a left leading to point \textbf{H} (17.439341, 78.363033), an underpass beneath the flyover.
    \item They choose the left route at \textbf{G}, leading to \textbf{H} beneath the flyover, constituting a broad U-turn with complex traffic dynamics.
    \item Retrace the route from \textbf{H} to \textbf{D}, and then conclude the journey at point \textbf{A}, contingent on a green signal.
    % \item Conclude the journey at point \textbf{A}, contingent on a green signal.
\end{enumerate}

The Tobii Glasses 2 records a video from its scene camera, processed using the Raw Gaze Filter in Tobii Analyzer software version 1.25. Each sampled frame is amalgamated to construct the gaze dataset.

The dataset comprises a total of 261,073 frames, with 150,767 frames from novice participants and 110,306 frames from experienced participants. Each frame features a distinctive red-colored circle, 0.5 cm in radius, indicating the participant's gaze location, serving as the gaze 'ground truth' for developing saliency models. The radius was fixed at 0.5 cm 
(minimum allowed) by the Tobii Analyzer software. Faces and vehicle license plates in the video recordings were blurred for anonymity using \textit{Understand AI'}s anonymiser, an open-source deep learning-based model \cite{anonym24}.

The data will be released in two formats to cater to different research needs. The first includes raw videos from the Tobii Glasses 2 scene camera, with faces of road users and number plates blurred to ensure anonymity. This version provides an ego-centric visual gaze on the dynamic real-world traffic scene data from two-wheeler drivers' perspective. The second format is preprocessed and consists of frames with gaze overlay in each frame (Figure \ref{fig2}c), extracted by removing frames without gaze data from the raw videos. This format enables researchers to effectively analyse gaze patterns and their relationship with the traffic environment. By providing both forms, researchers can run customized analyses to extract details on the objects on the road, the visual continuity maps from scan path analysis, and explore diverse aspects of driver behaviour.

\begin{figure} \centering
    \centering
    \includegraphics[width=0.8\linewidth]{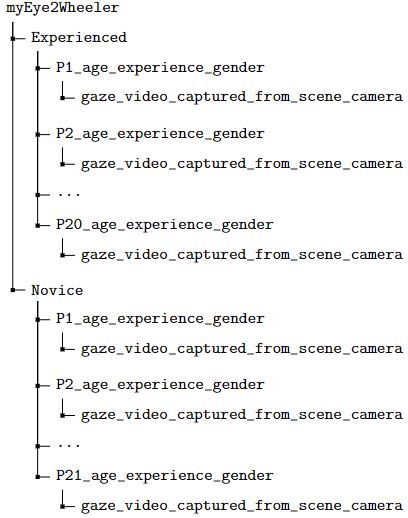}
    \caption{The data will be classified into Experienced and Novice, and each participant's age, two-wheeler driving experience and gender will provided in the folder's name containing that respective participant's data.}
    \label{tree}
\end{figure}

The \textit{myEye2Wheeler} dataset is organized into a hierarchical folder structure to facilitate easy access and analysis. Participants are categorized based on experience level (Experienced or Novice). Within these categories, individual participants are identified by \textit{Pn\_age\_experience\_gender}. Each participant's folder contains videos and data capturing the gaze direction and the scene viewed by the participant during the driving sessions. The structure is shown in detail in Figure \ref{tree}.

To enhance reproducibility and robustness and as recommended by reviewers, future studies can adopt a consistent train/test split methodology. That is, partition the dataset using a standardized approach whereby each participant's collected frames are divided sequentially. The initial 80\% of the frames obtained during the journey should constitute the training set, while the remaining 20\% should form the test set.

\begin{figure} \centering
    \centering
    \includegraphics[width=1\linewidth]{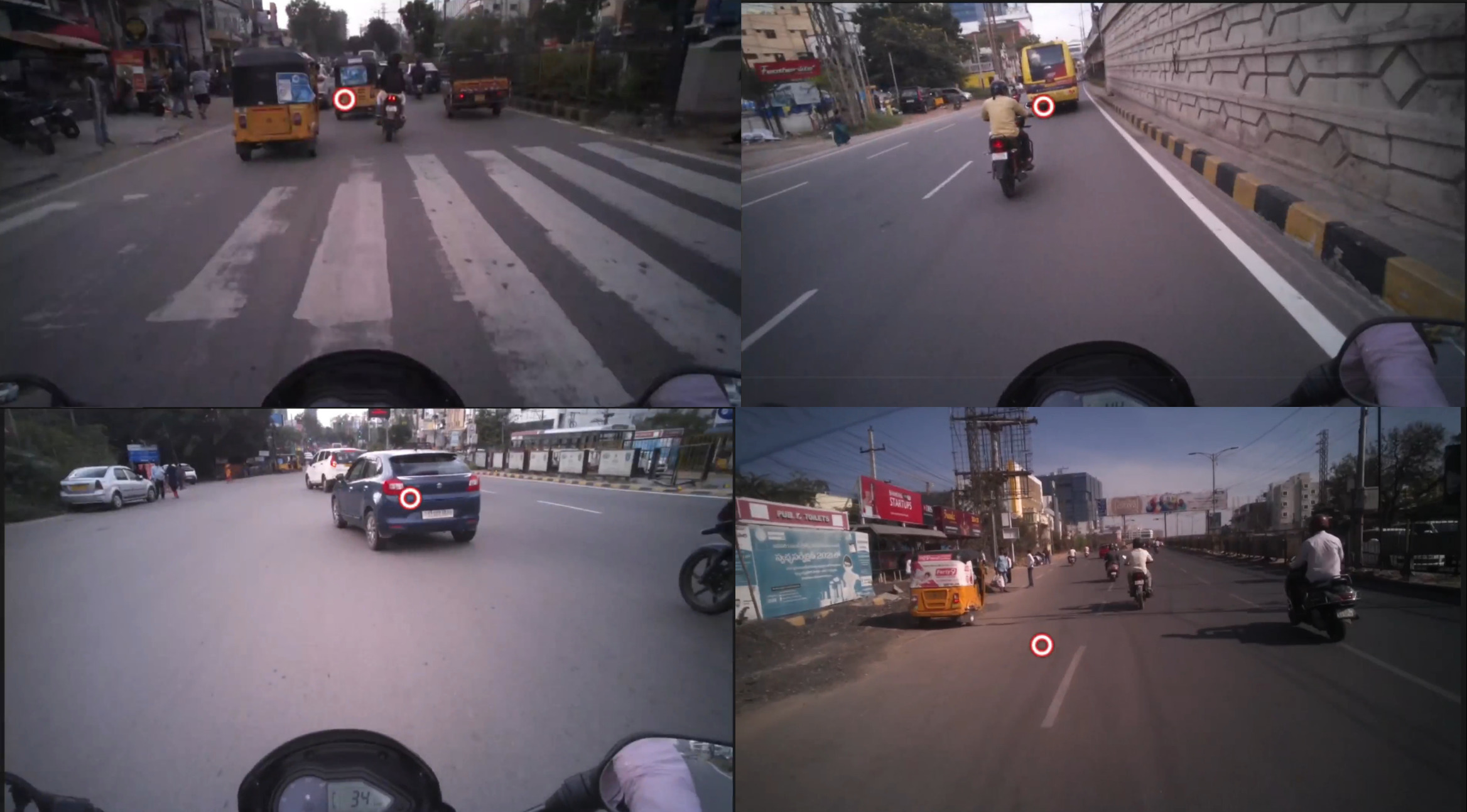}
    \caption{Some frames from the dataset}
    \label{frames}
\end{figure}

\section{COMPARATIVE ANALYSIS} 
A significant contribution of this dataset is to facilitate the development of saliency models tailored to Indian traffic scenarios. As a first attempt, we validate the efficacy of the newly curated myEye2Wheeler dataset against the established DR(Eye)VE dataset \cite{alletto2016dr6}, an extensive eye-tracking dataset from Europe. This validation also aims to assess the performance of the TASED-Net \cite{min201925} saliency model on both datasets to determine gaze prediction accuracy with ground truth (actual human gaze and field of view). A more complex and detailed model is being developed. 

TASED-Net, proposed in 2019, is a 3D fully convolutional network architecture for video saliency detection. It comprises two building blocks: the encoder network extracts low-resolution spatiotemporal features from input clips of multiple frames. In contrast, the prediction network spatially decodes the encoded features while temporally aggregating them. TASED-Net's temporal awareness, motion information processing, capacity for complex scene understanding, adaptability to heterogeneous data, and relevance to decision-making dynamics made it a promising saliency extraction model.

It is important to note that the model was initially trained on the DHF1K \cite{wang201826} dataset, consisting of random images with associated gaze point data. This step was applied to assess the effectiveness of general saliency data on driving data.

\subsection{Ground Truth Establishment} Yolov5 version v6.2 \cite{jocher202027} was used to classify objects in each frame belonging to both the datasets into 3 Areas of Interest(AOI): \textit{vehicles, pedestrians, on/off road}. These three AOIs were chosen as they represent critical elements in traffic surveillance and safety, addressing both vehicle movement and interaction with pedestrians while accounting for road conditions, aligning with the research objectives. The AOI, which corresponded to the gaze coordinates, became the ground truth gaze target. We are cognizant that further classification of the vehicles (bus, car, 2-wheeler, etc.) and lane estimation is required. As lane discipline is not highly adhered, we usually see five vehicles in a row in a 3-lane, for example. Hence, the detection of lane demarcation markers was not considered.

\subsection{Saliency Model Evaluation} 
The TASED-Net, trained on the DHF1K dataset, was employed as a saliency model to discern the most salient regions within each frame. The AOI with which the centroid of the identified salient region corresponded was determined as the predicted gaze targets. Accuracy of prediction was established by evaluating whether the class predicted by the TASED-Net aligned with the ground truth class for each frame. This comprehensive evaluation ensured the saliency model's reliability and precision in predicting driver gaze target classes. 
It was observed that TASED-Net accurately predicted the correct Gaze Target class in 84\% of the frames with the DR(Eye)VE dataset while achieving a lower rate of 61\% for the myEye2Wheeler dataset.
As an example to analyse the difference (Figure \ref{fig:enter-tdnet}), it is possible that the driver was looking at the gap between the motorcycle and the car estimate possible route, and thus the true gaze target here is \textit{on/off the road}, but the most salient part of the image predicted by TASED-Net is a vehicle. It is also possible that in frames prior to the current, the human driver has estimated the spatiotemporal location of the other vehicles based on experience and hence does not require constant attention. 

\begin{figure} \centering
    \centering
    \includegraphics[width=1\linewidth]{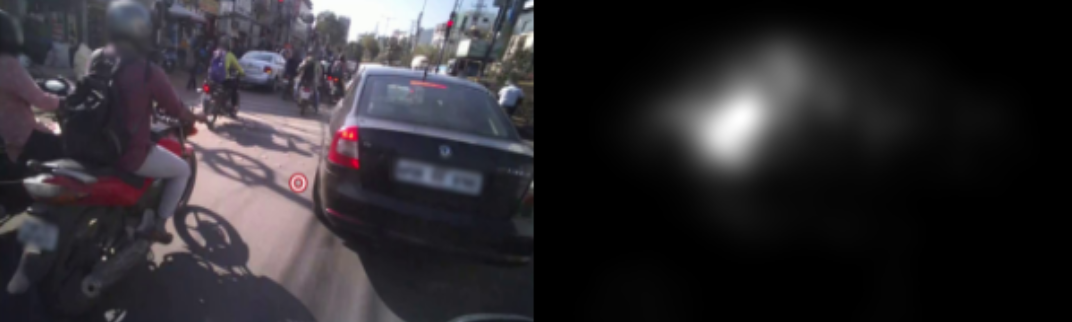}
    \caption{(a) Ground Truth (On/off-road) vs (b) Prediction (Vehicle in this case) }
    \label{fig:enter-tdnet}
\end{figure}

The variance observed in the TASED-Net's performance can be attributed to several factors, notably the dissimilarities between the training data and the conditions prevalent in the Indian traffic landscape. While the TASED-Net model was trained on the DHF1K dataset, which is not traffic-related but encompasses diverse video content, its performance on \textit{DR(Eye)VE} and \textit{myEye2Wheeler} gives interesting insights. The DR(Eye)VE dataset, collected in regions with comparatively less dense traffic and more adherence to structured traffic behaviour, allows for more accurate gaze prediction by models trained on similar norms. Conversely, \textit{myEye2Wheeler }dataset depicts the heterogeneity of Indian traffic, characterised by diverse driving behaviours, lane deviations, frequent gap exploitation, particularly prevalent among two-wheeler riders, and complex traffic dynamics, which could have led to lower accuracy. 
\begin{figure} \centering
    \centering
    \includegraphics[width=0.55\linewidth]{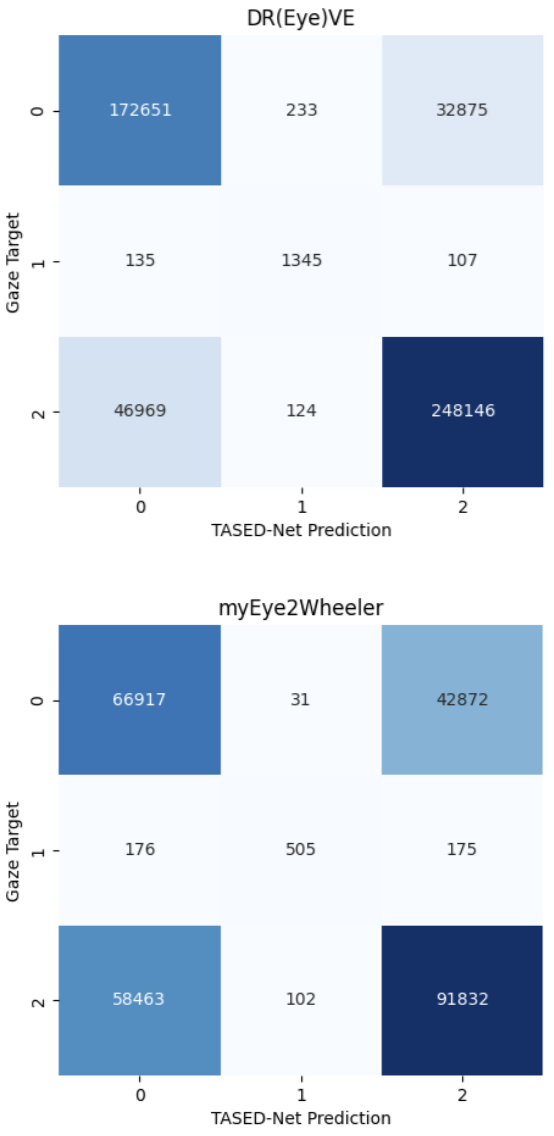}
    \caption{0, 1, 2 refer to "vehicles, "person", "on/off road" respectively. The confusion matrix indicates that TASED-Net exhibits a higher error rate for predictions, particularly when the actual gaze target is 0 "vehicles" but it tends to predict "on/off road" and vice versa, in both cases, however, it is significantly more in myEye2Wheeler's case.}
    \label{fig:enter-tdnet}
\end{figure}

\section{DISCUSSION} 
The \textit{myEye2Wheeler} dataset addresses several limitations observed in previous datasets focused on driver gaze behaviour. Unlike datasets such as DG-UNICAMP, DGW, DGAZE and DMD \cite{ribeiro16,ghosh17,Dua20,ortega18}, which primarily rely on stationary vehicle setups or interior observations, \textit{myEye2Wheeler} captures real-world driving scenarios on Indian roads, offering a more accurate representation of driving complexities, especially in dynamic traffic environments. Our dataset is similar to datasets such as LBW \cite{Isaac7} that also use wearable eye trackers but provide gaze data from four-wheeler drivers. It enables us to uncover nuances in gaze allocation dynamics in response to the evolving traffic landscape, facilitating a deeper understanding of cognitive processes that affect driving decisions. 

Furthermore, \textit{myEye2Wheeler} provides a diverse range of driving scenarios specific to Indian roads, unlike datasets like DADA-2000 and DR(eye)VE \cite{dada19,alletto2016dr6}. The dataset also offers information on accident estimations. This specificity allows researchers to study interactions with pedestrians, non-motorised vehicles, and vehicular traffic prevalent in Indian heterogeneous traffic conditions.

The diversity of traffic scenarios in myEye2Wheeler presents a significant challenge for TASED-Net, which is not explicitly tuned for Indian traffic conditions. This underscores the need for context-aware salience models tailored to dynamic and sometimes chaotic traffic scenarios. We are currently developing a specialised model to enhance gaze prediction accuracy in various traffic situations.

Focusing on two-wheeler drivers enriches discussions on traffic safety, particularly in regions of the world where the most cost-effective and fuel-efficient daily mode of transport is the 2-wheeler. Given that a city cannot expand lanes or roads to accommodate four-wheelers, which would be an environmental disaster, we highlight the need to foster more inclusive strategies, enhance road safety and advocate for contextually relevant approaches to traffic safety and driver behaviour research by considering this segment of vehicles.

\subsection{Conclusion}
By elucidating the factors influencing gaze behaviour and highlighting the importance of context-specific models in understanding driver behaviour, this dataset lays the groundwork for future advancements in road safety and human-machine interaction, particularly in regions with unique driving conditions such as India.

\subsection{Limitations}
% Technical challenges associated with wearable eye trackers, such as data loss due to connectivity issues and limitations in peripheral vision tracking, were observed. Addressing these issues could enhance the accuracy and reliability of eye-tracking data collection methods, contributing to a more comprehensive understanding of visual attention patterns during driving tasks.

The dataset for this study was collected during a narrow window from 2-5 PM, under uniform lighting conditions. This window allows for data capture when traffic flow is optimal, as peak-hours sees traffic jams. Driving data in mesopic or scoptic conditions requires different analysis methods, which will attempted in future datasets. 

The participant gender distribution is skewed (more male), which would affect behavioural analysis. Though, in experimental data collection all drivers exercise utmost caution while driving, making the behavior gender independent. An extension is to look at gig drivers as they employ interesting strategies to reach their destination.  In two-wheelers, we used a scooter
(lowest engine capacity model), while a motorbike (500cc)
can give us differential driver behaviour.
Comparative analysis was done with naturalistic datasets (like DR(eye)VE), and extending it to DGaze collected in lab conditions would give differential saliency networks, which is currently being explored. 

% The data was collected using a fixed, predetermined route, which does not account for the variability in road types, traffic density, and other environmental conditions. This limitation could affect the dataset's applicability to different traffic scenarios and limit the understanding of driver behavior under varied urban and rural road conditions. The specific predetermined route was chosen because we received permission from the traffic police commission to conduct the experiment on this route.

% The dataset evaluation primarily utilizes TASED-Net and a classification accuracy metric, which might not fully capture the nuanced and dynamic gaze behaviors typical of two-wheeler drivers in chaotic Indian traffic conditions. Incorporating additional models and metrics, such as confusion matrices and angular error measurements, could provide a more detailed understanding of gaze accuracy and its implications on safety.

% Data collection was restricted to non-peak hours for the safety of the participants and the pillion riders. Including data from peak traffic hours could offer valuable insights into driver behavior under more stressful and complex traffic conditions, which are currently not represented in the dataset. Collecting data during these periods would enhance the comprehensiveness of the study and increase its relevance to real-world driving scenarios.

\section{ACKNOWLEDGEMENT}
The authors would like to sincerely thank IHub-DATA, IIIT Hyderabad, for partial financial assistance. A special acknowledgement to the participants who volunteered for the study on working/office days.

\end{document}